\documentclass[journal]{IEEEtran}
\usepackage{graphicx}

\usepackage{cite}
\usepackage{xcolor}
\usepackage{color}

\ifCLASSINFOpdf
\else
\fi
\usepackage{amsmath}
\usepackage{framed,multirow}

\usepackage{algorithm}
\usepackage{algorithmic}
\begin{document}
%
% paper title
% Titles are generally capitalized except for words such as a, an, and, as,
% at, but, by, for, in, nor, of, on, or, the, to and up, which are usually
% not capitalized unless they are the first or last word of the title.
% Linebreaks \\ can be used within to get better formatting as desired.
% Do not put math or special symbols in the title.
\title{Decompose to Adapt: Cross-domain Object Detection via Feature Disentanglement}
%
%
% author names and IEEE memberships
% note positions of commas and nonbreaking spaces ( ~ ) LaTeX will not break
% a structure at a ~ so this keeps an author's name from being broken across
% two lines.
% use \thanks{} to gain access to the first footnote area
% a separate \thanks must be used for each paragraph as LaTeX2e's \thanks
% was not built to handle multiple paragraphs
%

\author{Dongnan~Liu,~\IEEEmembership{Member,~IEEE,}
        Chaoyi~Zhang, %~\IEEEmembership{Student member,~IEEE,}
        Yang~Song,~\IEEEmembership{Member,~IEEE,}
        Heng~Huang,~Chenyu Wang,~\\Michael Barnett
        and~Weidong~Cai,~\IEEEmembership{Member,~IEEE}% <-this % stops a space
\thanks{D. Liu, C. Zhang, and W. Cai are with the School of Computer Science, University of Sydney, NSW,
2008 Australia. Corresponding author: Dongnan Liu (dongnan.liu@sydney.edu.au)}% <-this % stops a space
\thanks{Y. Song is with the School of Computer Science and Engineering, University of New South Wales, NSW, 2052 Australia.}% <-this % stops a space
\thanks{H. Huang is with the Department of Electrical and Computer Engineering, University of Pittsburgh, Pittsburgh, PA 15261 USA.}
\thanks{C.~Wang and M. Barnett are with the Brain and Mind Centre, University of Sydney, NSW, 2050 Australia.}}

\maketitle

% As a general rule, do not put math, special symbols or citations
% in the abstract or keywords.
\begin{abstract}

Recent advances in unsupervised domain adaptation (UDA) techniques have witnessed great success in cross-domain computer vision tasks, enhancing the generalization ability of data-driven deep learning architectures by bridging the domain distribution gaps. For the UDA-based cross-domain object detection methods, the majority of them alleviate the domain bias by inducing the domain-invariant feature generation via adversarial learning strategy. However, their domain discriminators have limited classification ability due to the unstable adversarial training process. Therefore, the extracted features induced by them cannot be perfectly domain-invariant and still contain domain-private factors, bringing obstacles to further alleviate the cross-domain discrepancy. To tackle this issue, we design a Domain Disentanglement Faster-RCNN (DDF) to eliminate the source-specific information in the features for detection task learning. Our DDF method facilitates the feature disentanglement at the global and local stages, with a Global Triplet Disentanglement (GTD) module and an Instance Similarity Disentanglement (ISD) module, respectively. By outperforming state-of-the-art methods on four benchmark UDA object detection tasks, our DDF method is demonstrated to be effective with wide applicability.

\end{abstract}

% Note that keywords are not normally used for peerreview papers.

% For peer review papers, you can put extra information on the cover
% page as needed:
% \ifCLASSOPTIONpeerreview
% \begin{center} \bfseries EDICS Category: 3-BBND \end{center}
% \fi
%
% For peerreview papers, this IEEEtran command inserts a page break and
% creates the second title. It will be ignored for other modes.
\IEEEpeerreviewmaketitle

\section{Introduction}

\IEEEPARstart{O}{bject} detection, which aims to assign a bounding box and category prediction for each foreground instance, is essential for modern computer vision. Taking advantages from the deep learning techniques, previous object detection methods based on convolutional neural networks (CNN) have achieved appealing performance on various benchmarks~\cite{girshick2015fast,ren2015faster,redmon2016you,lin2017focal,cai2018cascade}. However, these fully-supervised models have been criticized for the lack of generalization ability and suffer from severe performance drop when validated on other unseen datasets, since they tend to bias towards the data distribution of the training domain~\cite{torralba2011unbiased,yosinski2014transferable}. On the other hand, collecting sufficient annotations for each new domain is impractical in real applications, due to time-consuming and expensive annotation procedure.

\begin{figure}[htbp]
\centering
\includegraphics[width=0.49\textwidth]{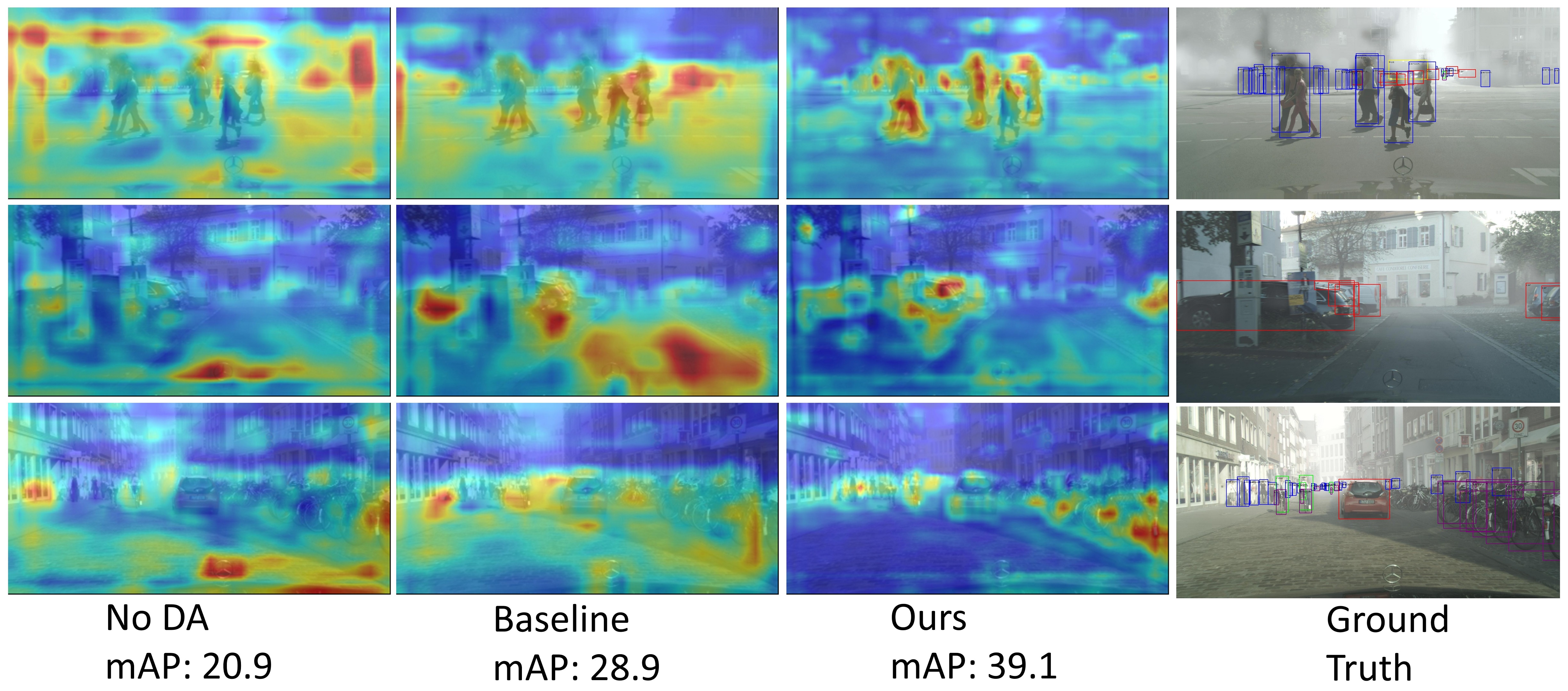}

\caption{{\color{black}{The visualization results on the image-level features. `No DA' is the Faster R-CNN without adaptation, and `Baseline' is the Faster R-CNN with basic adversarial domain discriminator. For the No DA and Baseline, the features are collected and visualized from the backbone network. For our method, we simultaneously visualize the global-level domain-shared and domain-private features. All the features and labels are overlapped on the original images.~\textbf{(Best viewed in color and zoom in)}}} }
\label{fig-disent-c2f-intro}
\end{figure}

To address this dilemma, unsupervised domain adaptation (UDA) methods have been proposed to transfer the domain-invariant information from the labeled source domain to an unlabeled target domain~\cite{pan2009survey,ganin2014unsupervised,tzeng2017adversarial}. For the UDA object detection methods, the majority of them incorporate the adversarial domain discriminators with the object detection architectures (e.g Faster-RCNN), to alleviate the cross-domain discrepancy by inducing the domain-invariant features generation~\cite{chen2018domain,he2019multi,saito2019strong,he2020domain}.

{\color{black}{
Despite the state-of-the-art performance of the adversarial learning-based UDA object detection methods, they overlook the entanglement between the $domain$-$shared$ and $domain$-$private$ features in the latent manifold space~\cite{bousmalis2016domain,cai2019learning,peng2019domain}. Due to the unstable learning process, the decision boundary for the adversarial domain classifiers in these methods is inaccurate~\cite{arjovsky2017wasserstein}. As such, the extracted features in the typical adversarial UDA methods cannot be perfectly domain-invariant and inevitably contain domain-private factors. Optimizing the UDA models with these entangled features induces the model bias towards the source domain, and further degrades the performance on the target domain. As indicated in Figure~\ref{fig-disent-c2f-intro}, the Baseline method with the adversarial domain discriminator tends to focus on the source-specific information in the background, which leads to suboptimal adaptation performance.
}}

Previously, several methods have been proposed to solve the feature entanglement issue in UDA classification~\cite{bousmalis2016domain,cai2019learning,peng2019domain,peng2020domain2vec,gholami2020unsupervised}. They are established based on auto-encoder architectures, incorporated with latent code independence induction mechanism, and classifier entropy regularization to decompose the domain-private factors from the features for classifier learning. However, directly extending the feature disentanglement-based UDA classification methods to cross-domain object detection suffers from challenges. First,~\cite{bousmalis2016domain,cai2019learning,peng2019domain} decompose the shared and private features in each domain by inducing their distributions to be independent, which always requires the batch size of the features to be sufficiently large to describe the characteristics of the distribution. However, the batch size for UDA object detection is very limited (sometimes even equals to 1~\cite{saito2019strong,chen2018domain}),  due to the mandatory larger model sizes than those for classification. Second, these methods ignore the feature entanglement at the local level for each foreground object since their objective is to assign a category label for the whole image. For cross-domain object detection which typically contains multiple object instances in an image, the adaptation ability of these UDA disentanglement-based classification model is limited, resulting from the instance-level feature entanglement. {\color{black}{Although PD~\cite{wu2021instance} was previously proposed for UDA object detection via feature decomposition based on mutual information minimization~\cite{peng2019domain}, the limited batch size for the object detection framework causes the suboptimal adaptation performance. Subsequently, there currently still lack feature disentanglement UDA methods particularly for cross-domain object detection. }}

Motivated by the aforementioned observations, we propose a Domain Disentanglement Faster-RCNN (DDF) method in this work, to improve the typical adversarial learning-based UDA object detection via feature disentanglement. Specifically, our DDF achieves feature disentanglement at the global and local levels. {\color{black}{Throughout this paper, the global-level features represent the output of the backbone network, which contain the information of the object structure, and spatial distribution for the whole images. The local-level features represent the region of interests (ROIs) for object localization and classification. Domain-invariant features represent the features containing shared factors between the source and target domains, and the instance-invariant features represent the domain-invariant features at the instance-level for the ROIs.}} First, we design a Global Triplet Disentanglement (GTD) module jointly optimized with a domain discriminator, which improves the feature adaptation ability at the global level via a triplet feature disentanglement strategy. To further facilitate the feature disentanglement at the local stage, an Instance Similarity Disentanglement (ISD) module is proposed, based on the similarity regularization between the shared and private features for instance objects. Our DDF method is validated on four benchmark UDA object detection tasks and outperforms the state-of-the-art methods.

\section{Related Work}

\subsection{Unsupervised Domain Adaptation}

Unsupervised domain adaptation (UDA) aims at bridging the gap between an annotated source domain and an unannotated target domain. Typically, UDA methods transfer the knowledge in {\color{black}{four ways}}: 1) directly minimizing the statistical distribution distance between two domains~\cite{gretton2007kernel,fernando2013unsupervised}; 2) inducing the domain-invariant feature generation~\cite{tzeng2017adversarial,ganin2014unsupervised,shermin2020adversarial}; 3) learning from the synthesized images~\cite{zhu2017unpaired,huang2018multimodal,hoffman2017cycada,emami2020spa,liu2020pdam}, and {\color{black}{4) self-training via pseudo labels~\cite{inoue2018cross,huang2021cross}}}. By alleviating the cross-domain discrepancy at the feature and appearance levels, UDA methods have achieved outstanding performance in cross-domain classification~\cite{ganin2014unsupervised,tzeng2017adversarial,hoffman2017cycada}, segmentation~\cite{li2019bidirectional,vu2019advent,liu2020unsupervised}, and detection~\cite{chen2018domain,saito2019strong,inoue2018cross,zhu2019adapting,chen2020harmonizing,he2020domain}. Since our proposed method aims at tackling the feature entanglement issue for cross-domain object detection, only the literature of UDA object detection and feature disentanglement are reviewed in detail.

\begin{figure*}[t]
\centering
\includegraphics[width=0.85\textwidth]{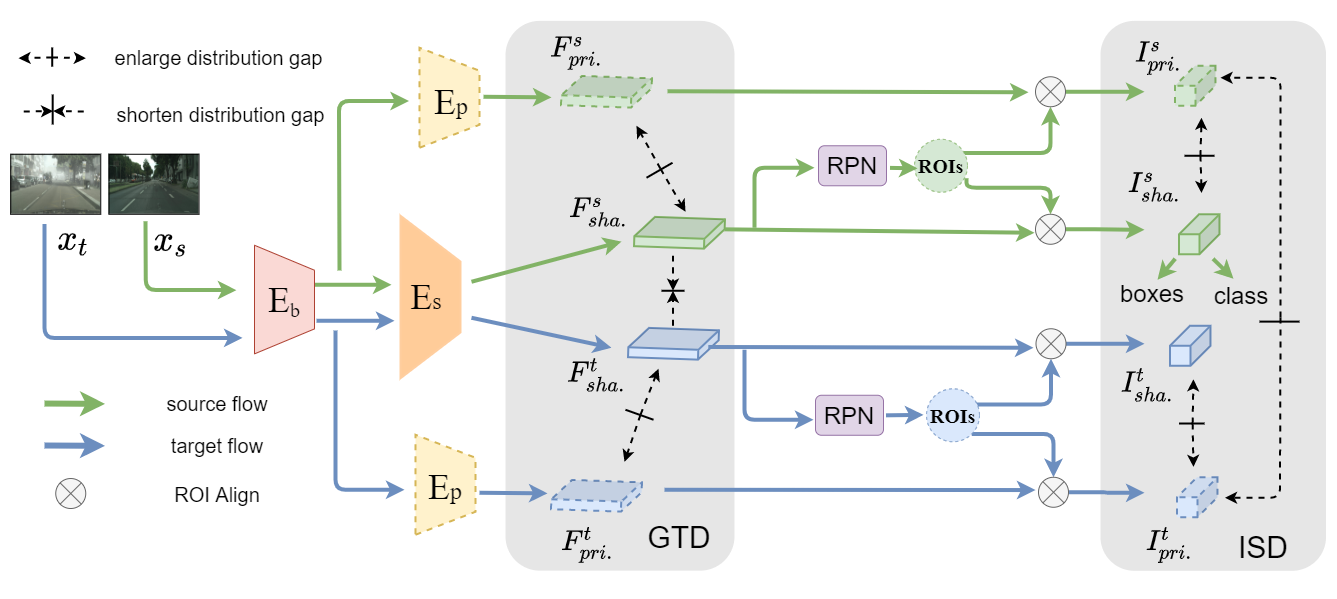} % Reduce the figure size so that it is slightly narrower than the column.
\caption{\color{black}{The overall architecture of our proposed DDF method. Please refer to Section~\ref{sec-overall} for detailed definitions on all the symbols. Note that the $MLP$ module before $I_{sha.}^{s}$, $I_{sha.}^{t}$, $I_{pri.}^{s}$ and $I_{pri.}^{t}$ is omitted.~\textbf{(Best viewed in color)}}}
\label{fig-overall}
\end{figure*}

\subsection{UDA for Cross-domain Object Detection}

Recent UDA object detection methods are designed based on the two-stage detector Faster-RCNN and induce the domain-invariant feature learning by domain adaptive modules at the feature and appearance levels. Among the feature-level adaptation methods, Chen~\emph{et al.}~\cite{chen2018domain} first proposed to learn domain-invariant features for the global image and local instance features. To align the image-level features with different scales, strong-weak feature alignment was then proposed~\cite{saito2019strong}.~\cite{zhu2019adapting} designed a selective adaptation architecture to focus on the local regions. Later, hierarchical feature alignment with multi-stage domain discriminators further improves the baseline~\cite{he2019multi,chen2020harmonizing,shen2020cdtd}. Additionally, category-aware feature alignment is effective in alleviating the class-imbalance issues during adaptation~\cite{xu2020exploring,zhuang2020ifan,xu2020cross,zheng2020cross}. {\color{black}{In~\cite{huang2021fsdr}, a domain randomization strategy is proposed based on the spectrum learning at the frequency space, which has achieved appealing performance on the cross-domain object detection under the domain generalization setting.}} For the appearance-level adaptation, target-like synthetic images and pseudo labels are commonly employed~\cite{inoue2018cross,kim2019diversify}. Even though the previous methods alleviated the domain-bias by various feature-level or appearance-level adaptations, few of them have discussed and attempted to explicitly address the entanglement of the domain-shared and domain-private factors, limiting the model's adaptation abilities.

\subsection{UDA with Feature Disentanglement}

Previous feature disentanglement methods decomposed the features by encouraging the independence between them, based on the variation autoencoders (VAE) \cite{kingma2013auto,higgins2017beta} and generative adversarial networks (GAN)~\cite{mathieu2016disentangling,chen2016infogan}. By separating the appearance-level (e.g., texture, brightness) and the global-level factors (e.g., object structure, spatial distribution) in the latent feature space,~\cite{huang2018multimodal,lee2018diverse} achieved compelling performance on multi-modal image synthesis. For UDA classification,~\cite{bousmalis2016domain,cai2019learning,peng2019domain,peng2020domain2vec} have been proposed to decompose the domain-private and domain-shared information using autoencoder-based structures. As essential steps for these methods, the independence induction for the latent feature distribution requires large batch sizes, which is impractical for UDA object detection. {\color{black}{In addition,~\cite{kurmi2019attending} proposed to decompose the features via the attention mechanism based on uncertainty learning. In~\cite{deng2021informative}, a variational information bottleneck was designed to filter out the redundant domain-private information and achieve the disentanglement.}} However, the feature entanglement issue for each object instance has not been considered among these methods, which is essential for UDA object detection. Therefore, it is not intuitive to directly extend current UDA feature disentanglement methods for cross-domain object detection. 

{\color{black}{In~\cite{su2020adapting}, a Conditional Domain Normalization (CDN) mechanism was proposed for UDA object detection via feature decomposition, by incorporating the source-specific factors with the target features. However, there lacks feature decomposition for the target features, which would cause the model to suffer from domain bias during inference.}} {\color{black}{Although Wu~\emph{et al.}~\cite{wu2021instance} designed a PD method with progressive feature disentanglement mechanism for UDA object detection, their feature entanglement modules are based on Mutual Information minimization~\cite{peng2019domain}, which requires batch-wise feature shuffling~\cite{belghazi2018mutual}. Based on the implementation of~\cite{peng2019domain}, the features have a batch size of $128$, sufficiently large to contain the distribution characteristics of their corresponding domains. When adapting the method to UDA object detection, the disentanglement module in~\cite{wu2021instance} has to work with small batch sizes and hence could not fully utilize the Mutual Information minimization strategy. To tackle this issue, we propose a DDF method for cross-domain object detection via feature disentanglement at the global and local levels, without requiring a large image batch size for training. Particularly, our DDF method has achieved better performance than PD~\cite{wu2021instance} (detection mAP: Ours: 39.1; PD: 36.6). This further demonstrates the superiority of our designed feature decomposition modules compared to the typical ones, when the batch sizes for the framework are limited for cross-domain object detection.}}

%-------------------------------------------------------------------------

\section{Domain Disentanglement Faster-RCNN}

In this section, we present the detailed framework of our Domain Disentanglement Faster-RCNN (DDF) model. Denote a labeled source domain with $N_{s}$ i.i.d images as $D_{s}=\{(x^{s}_{i}, c^{s}_{i}, b^{s}_{i})\}_{i = 1, 2,...,N_{s}}$, where $c^{s}_{i}$ and $ b^{s}_{i}$ represent the category labels and bounding box coordinates for all the foreground objects in each image $x^{s}_{i}$, respectively. Then, an unlabeled target domain with $N_{t}$ i.i.d images is defined as $D_{t}=\{(x^{t}_{j})\}_{j = 1, 2,...,N_{t}}$, with a different data distribution from the $D_{s}$. Our DDF method aims at transferring the knowledge from the labeled $D_{s}$ to the unlabeled $D_{t}$ and achieving competitive detection performance on the target domain.

\subsection{Framework Overview~\label{sec-overall}}

The overall diagram of our proposed DDF model is shown in Figure~\ref{fig-overall}. First, we establish a baseline UDA object detection model, by incorporating a global-level adversarial domain discriminator with a Faster-RCNN detection model. During each training iteration, the source and target images, defined as $x_{s}$ and $x_{t}$, are fed into the network. First, a backbone is employed to extract the global-level features of the input images, constructed by a basic feature encoder $E_{b}$ with fixed weights, and a domain-shared feature extractor $E_{s}$ with dynamically updated weights.

To facilitate the feature disentanglement, we design a domain-private feature encoder $E_{p}$ following the $E_{b}$, to obtain the feature maps containing specific factors of each domain. As shown in Figure~\ref{fig-overall} and Equation~\ref{equ-init-global}, $F_{sha.}^{s}$ and $F_{sha.}^{t}$ represent the domain-shared features across the source and target domains, and $F_{pri.}^{s}$ and $F_{pri.}^{t}$ indicate the domain-private features specific to the source and target domains:

% \small
\begin{equation}
\begin{aligned}
F_{sha.}^{s}= E_{s}(E_{b}(x_{s})), ~~~~~~   F_{sha.}^{t} = E_{s}(E_{b}(x_{t})),  \\
F_{pri.}^{s} = E_{p}(E_{b}(x_{s})),   ~~~~~~   F_{pri.}^{t} = E_{p}(E_{b}(x_{t})).
\label{equ-init-global}
\end{aligned}
\end{equation}
% \normalsize

For the global-level feature disentanglement, $F_{sha.}^{s}$, $F_{sha.}^{t}$, $F_{pri.}^{s}$, and $F_{pri.}^{t}$ are optimized with the Global Triplet Disentanglement (GTD) module, which aligns the distributions of the domain-shared features between the source and target domains, as well as enlarges the discrepancy between the domain-shared and domain-private features within each domain. Leveraging the GTD module, the global-level domain-specific factors are thus decomposed from the domain-shared features for the detection task learning. 

At the local level, we employ a Region Proposal Network (RPN) and a ROIAlign layer to extract the local instance features in $F_{sha.}^{s}$, $F_{sha.}^{t}$, $F_{pri.}^{s}$, and $F_{pri.}^{t}$:

% \small
\begin{equation}
\begin{aligned}
I_{sha.}^{s} & = MLP\Big(ROIAlign\big(F_{sha.}^{s},~RPN(F_{sha.}^{s})\big)\Big), \\
I_{sha.}^{t}& = MLP\Big(ROIAlign\big(F_{sha.}^{t},~RPN(F_{sha.}^{t})\big)\Big),  \\
I_{pri.}^{s} &= MLP\Big(ROIAlign\big(F_{pri.}^{s},~RPN(F_{pri.}^{s})\big)\Big),  \\
I_{pri.}^{t} &= MLP\Big(ROIAlign\big(F_{pri.}^{t},~RPN(F_{pri.}^{t})\big)\Big),
\label{equ-init-local}
\end{aligned}
\end{equation}
% \normalsize
where $MLP$ contains a flatten layer followed by three fully connected layers. In Equation~\ref{equ-init-local}, $I_{sha.}^{s}$ and $I_{sha.}^{t}$ denote the instance-wise domain-shared features captured from each domain, while $I_{pri.}^{s}$ and $I_{pri.}^{t}$ represent the domain-private ones. Then, the instance-level $I_{sha.}^{s}$, $I_{sha.}^{t}$, $I_{pri.}^{s}$, and $I_{pri.}^{t}$ are fed into the Instance Similarity Disentanglement (ISD) module to facilitate the feature disentanglement at the local stage, based on feature similarity optimization. Eventually, the domain-shared features at the instance-level ($I_{sha.}^{s}$) are employed for the object classification and location regression learning. The detailed structure of the private components of the domain-specific feature extractor is shown in Table~\ref{table-overall-net}.

\begin{table}[!htb]
\centering
\caption{Detailed model structure of the overall architecture. $in$ and $out$ represent the input and output channel numbers of each layer, respectively. $k$, $s$, and $p$ are the kernel size, stride, and padding of each layer.}
\resizebox{0.97\linewidth}{!}{%
\begin{tabular}{c|l|l}
\hline
Layer  & $E_{p}$ & Hyperparameters  \\
\hline
1  & Conv, ReLU & $k = (3, 3), s = 1, p = 1, in = 256, out = 256$  \\
2 & Maxpooling & $k = (2, 2), s = 2, p = 0$  \\
3  & Conv, ReLU & $ k = (3, 3), s = 1, p = 1, in = 256, out = 512$  \\
4 & Maxpooling & $k = (2, 2), s = 2, p = 0$  \\
5  & Conv, ReLU & $k = (3, 3), s = 1, p = 1, in = 512, out = 512$  \\
\hline
\hline
Layer  & $D_{glb}$ & Hyperparameters  \\
\hline
1  & Conv, ReLU & $k = (3, 3), s = 1, p = 1, in = 512, out = 512$  \\
2  & Conv, ReLU & $k = (3, 3), s = 1, p = 1, in = 512, out = 256$  \\
3  & Conv, ReLU & $k = (3, 3), s = 1, p = 1, in = 256, out = 128$  \\
4  & Conv, ReLU & $k = (3, 3), s = 1, p = 1, in = 128, out = 64$  \\
5  & Conv & $ k = (3, 3), s = 1, p = 1, in = 64, out = 2$  \\
\hline
\end{tabular}}
\label{table-overall-net}
\end{table}

\subsection{Global Triplet Disentanglement\label{sec-gtd}}

\begin{figure}[t]
\centering
\includegraphics[width=0.3\textwidth]{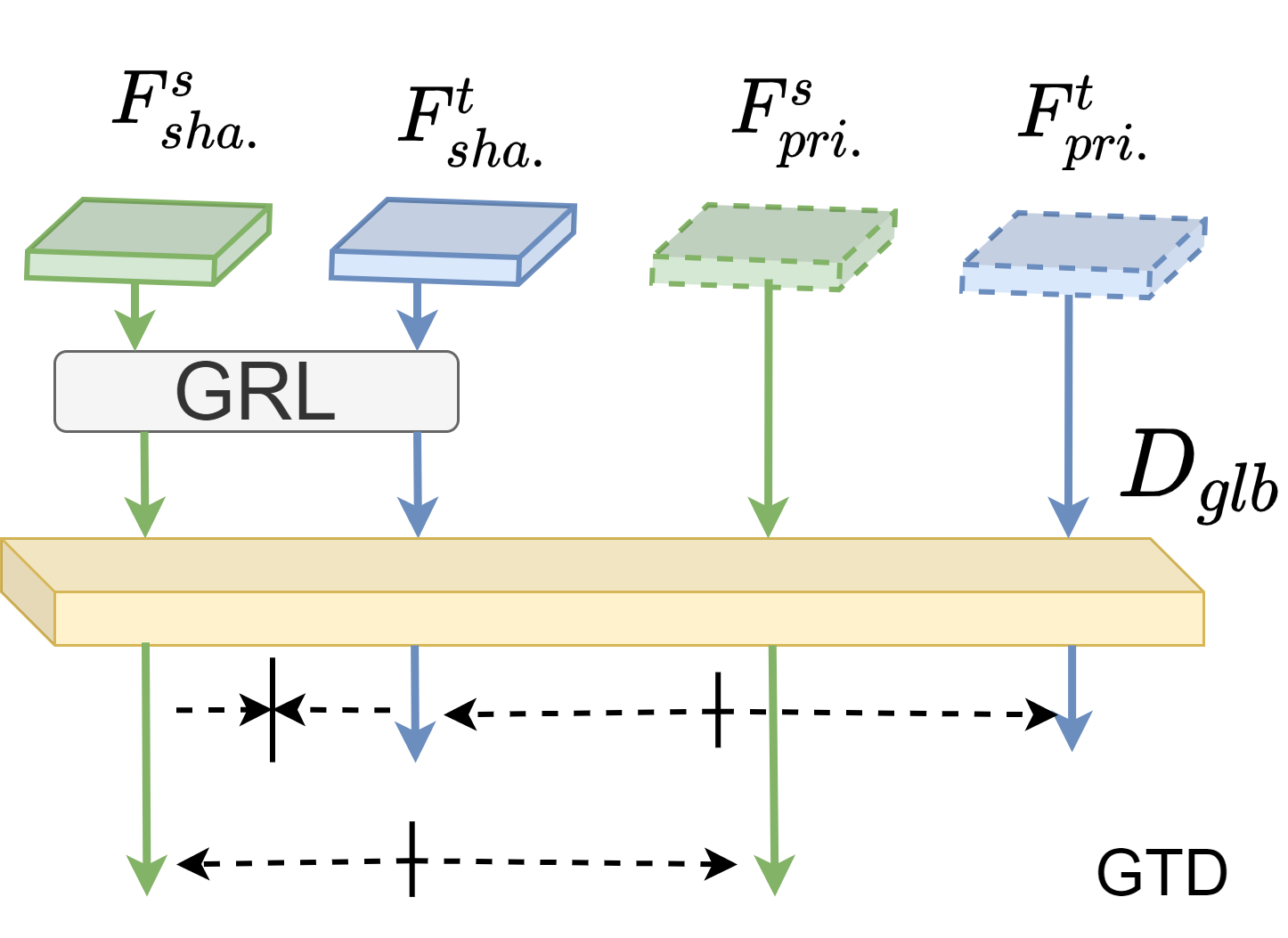} % Reduce the figure size so that it is slightly narrower than the column.
\caption{The detailed structure of our proposed Global Triplet Disentanglement (GTD) module in Section~\ref{sec-gtd}. The definitions for all the symbols are the same as Figure~\ref{fig-overall}.~\textbf{(Best viewed in color, and in conjunction with Figure~\ref{fig-overall})}}
\label{fig-gtd}
\end{figure}

In previous UDA object detection methods, the invariance between the domain-shared $F_{sha.}^{s}$ and $F_{sha.}^{t}$ is induced by optimizing an adversarial domain discriminator at the global level~\cite{chen2018domain,saito2019strong,he2020domain} through:
% \small
\begin{equation}
\begin{aligned}
L_{di} & = \max\limits_{\theta_{E_{s}}} \min\limits_{\theta_{D}} L_{ce}\Big(D_{glb}(F_{sha.}^{s}), 0\Big) + L_{ce}\Big(D_{glb}(F_{sha.}^{t}), 1\Big),
\label{equ-loss-di}
\end{aligned}
\end{equation}
% \normalsize
where $L_{ce}(\cdot, \cdot)$ measures the cross entropy between the feature maps and labels, while $\theta_{E_{s}}$ and $\theta_{D}$ indicate the parameters for the domain-shared feature extractor $E_{s}$ and domain discriminator $D_{glb}$, respectively. We denote the domain label as $0$ for the source domain and $1$ for the target domain. However, as suggested in~\cite{arjovsky2017wasserstein,cai2019learning}, obtaining the precise classification decision boundaries for the adversarial discriminators in typical UDA methods is always challenging due to the unstable learning procedure. As such, the features induced by $D_{glb}$ in Equation~\ref{equ-loss-di} cannot be perfectly domain-invariant, and some lurking domain-private factors would inevitably bring domain bias into the detection task learning. {\color{black}{Although recent autoencoder-based feature disentanglement methods attempt to alleviate this issue for UDA classification tasks~\cite{bousmalis2016domain,cai2019learning,peng2019domain}, they are suboptimal for UDA object detection architectures. Due to the large model size, the batch size for the typical UDA object detection framework is always set to 1~\cite{chen2018domain,saito2019strong,wu2021instance}. However, the feature independence learning paradigms for feature disentanglement in~\cite{bousmalis2016domain,cai2019learning,peng2019domain} involve the calculations across a batch with average and shuffling, which would become redundant when the batch size is 1.}}

{\color{black}{To this end, we propose a Global Triplet Disentanglement (GTD) module, which aims to improve the adversarial learning-based feature adaptation at the global level via feature disentanglement. Our GTD module focuses on the triplet optimization among the domain-private and domain-shared features across two domains without batch-wise calculations as~\cite{bousmalis2016domain,cai2019learning,peng2019domain}, hence mitigating the influence of feature batch size.}} Detailed illustrations of the GTD module are shown in Figure~\ref{fig-gtd}. {\color{black}{We consider that under the ideal situation, the domain-private $F_{pri.}^{s}$ and $F_{pri.}^{t}$ contain the specific factors for the source and target domains, respectively.}} Subsequently, the domain discriminator $D_{glb}$ should be able to distinguish them. Based on this assumption, we first introduce a domain-specific classification loss $L_{ds}$ to enhance the classification ability of $D_{glb}$:
% \small
\begin{equation}
\begin{aligned}
L_{ds} & =  \min\limits_{\theta_{D}, \theta_{E_{p}}} L_{ce}\Big(D_{glb}(F_{pri.}^{s}), 0\Big) + L_{ce}\Big(D_{glb}(F_{pri.}^{t}), 1\Big),
\label{equ-loss-ds}
\end{aligned}
\end{equation}
% \normalsize
where $\theta_{E_{p}}$ indicates the parameters of the domain-private encoder $E_{p}$.

To further enlarge the discrepancy between the domain-shared and private features within each domain, and meanwhile to align the domain-shared features across the domains, we propose a triplet optimization strategy for $F_{sha.}^{s}$, $F_{sha.}^{t}$, $F_{pri.}^{s}$, and $F_{pri.}^{t}$. Specifically, the triplet feature decomposition loss $L_{tri}$ is defined as:
% \small
\begin{equation}
\begin{aligned}
L_{tri} & =\frac{1}{2}\bigg(\max\Big(d(F_{sha.}^{s}, F_{sha.}^{t}) - d(F_{sha.}^{s}, F_{pri.}^{s}) + m,~0\Big) \\
& + \max\Big(d(F_{sha.}^{s}, F_{sha.}^{t}) - d(F_{sha.}^{t}, F_{pri.}^{t}) + m,~0\Big)\bigg),
\label{equ-loss-tri}
\end{aligned}
\end{equation}
% \normalsize
where $d(f_{1}, f_{2}) = ||D_{glb}(f_{1}) - D_{glb}(f_{2})||^{2}$, measuring the square Euclidean distance between the $SoftMax$ domain predictions of $f_{1}$ and $f_{2}$. We have set the margin value $m$ as $0.25$ to ensure the domain probability predictions of the $F_{sha.}^{s}$ and $F_{sha.}^{t}$ close to $0.5$. {\color{black}{In Equation~\ref{equ-loss-di}, $D_{glb}$ for distinguishing $F^s_{sha}$ from $F^t_{sha}$ is trained in an adversarial manner with the feature extractor $E_{s}$. Under the ideal optimization scenario, the features from $E_{s}$ should be able to confuse $D_{glb}$ for being domain-invariant. To this end, the domain category predictions for both $F^s_{sha}$ and $F^t_{sha}$ are expected to be $0.5$, which is consistent with the objective of Equation~\ref{equ-loss-tri}.}}

Overall, the GTD module improves the adaptation ability of the adversarial domain discriminator by decomposing the global-level source-specific factors. Meanwhile, the GTD module could be intuitively implemented without any batch-wise calculations, which is more practical-friendly to UDA object detection tasks by loosening the restrictions on large batch size. The loss function for the GTD module is defined as:
% \small
\begin{equation}
\begin{aligned}
L_{GTD} & =  L_{ds} + L_{tri}.
\label{equ-GTD}
\end{aligned}
\end{equation}
% \normalsize

\subsection{Instance Similarity Disentanglement}

The GTD module promotes the global-level feature adaptation by removing the domain-specific factors. However, there still exist domain-bias issues in the local-level features of each object. In previous work, adversarial domain discriminators are employed to alleviate the discrepancy for the cross-domain instance features at the local levels~\cite{chen2018domain,he2019multi}. However, similar to the global-level issues, the local-level adversarial learning-based feature adaptation also suffers from the feature entanglement, due to its inaccurate decision boundary for domain classification. 

To this end, we propose an Instance Similarity Disentanglement (ISD) to facilitate the instance-level feature alignment based on feature similarity optimization. First, we obtain the domain-shared and domain-private features at the local level, following Equation~\ref{equ-init-local}. Since the batch sizes of the local-level $I_{sha.}^{s}$, $I_{sha.}^{t}$, $I_{pri.}^{s}$ and $I_{pri.}^{t}$ features equal to the number of ROIs, they are sufficiently large for dependency-aware optimization. Based on the assumption that the domain-shared factors in each domain should be independent from the domain-private ones~\cite{bousmalis2016domain,cai2019learning,peng2019domain}, we next propose to enlarge the distribution distance between the domain-shared and domain-private instance features within each domain. This step could be achieved by minimizing their cosine similarity as:

% \small
\begin{equation}
\begin{aligned}
L_{ISD-intra} & = sim(I_{sha.}^{s}, I_{pri.}^{s}) + sim(I_{sha.}^{t},I_{pri.}^{t}),
\label{equ-ins-intra}
\end{aligned}
\end{equation}
% \normalsize
where $sim (\cdot, \cdot)$ denotes the cosine similarity between two vectors: $sim(a, b)  = \frac{a^T b}{||a||~||b||}$. Since $I_{sha.}^{s}$, $I_{sha.}^{t}$, $I_{pri.}^{s}$ and $I_{pri.}^{t}$ are acquired after the $ReLU$ activation layers, the cosine similarity between each pair of them is in range $[0, 1]$. In addition to the intra-domain feature similarity minimization in Equation~\ref{equ-ins-intra}, we then propose to minimize the feature similarity between the domain-private features across two domains. This is motivated by~\cite{chattopadhyay2020learning}: under the ideal disentanglement scenario, each pair of domain-private factors in the manifold space from two different domains should not intersect. Therefore, enlarging the distribution distance between the domain-private $I_{pri.}^{s}$ and $I_{pri.}^{t}$ is beneficial to facilitate the instance-level feature disentanglement, which could be mathematically described as minimizing:

% \small
\begin{equation}
\begin{aligned}
L_{ISD-inter} & = sim(I_{pri.}^{s},I_{pri.}^{t}).
\label{equ-ins-inter}
\end{aligned}
\end{equation}
% \normalsize

The loss function for the ISD module is defined by integrating the intra- and inter-domain feature disentanglement at the local stage:
% \small
\begin{equation}
\begin{aligned}
L_{ISD} & = L_{ISD-intra} + L_{ISD-inter}. 
        % & = sim(I_{ss}, I_{ps}) + sim(I_{st}, I_{pt}) + sim(I_{ps}, I_{pt})
\label{equ-isd}
\end{aligned}
\end{equation}
% \normalsize
% Note that the ISD module is optimized without adversarial learning.

\subsection{Training and Inference\label{sec-imp}}

Our DDF method is build on Faster-RCNN with a VGG16 and ResNet50 backbone. For model initialization, the backbones are pretrained on the ImageNet classification task~\cite{deng2009imagenet} and the weights of the remaining layers are initialized with normal distribution initialization. Following the implementation of~\cite{saito2019strong,chen2020harmonizing,xu2020cross}, the fixed-weight $E_{b}$ is the model before the $conv3\_3$ layer of the VGG16 backbone, and $E_{s}$ is the rest. The batch size is $1$ and we do not use batch normalization due to the small mini-batch size. Each batch contains two images, one from the source domain and the other from the target domain. For all the experiments, the shorter side of the training image is resized to $600$, while maintaining the aspect ratio. During training, we employ the SGD optimizer with a momentum of $0.9$, and the weight decay is set to $0.0005$. The initial learning rate is $0.001$ for the first $50K$ training iterations and decreased to $0.0001$ for the last $20K$ iterations.

The overall loss function of our proposed DDF is defined as:

% \small
\begin{equation}
\begin{aligned}
L_{ddf} & = L_{det} + L_{di} + L_{GTD} + L_{ISD}, 
\label{equ-loss-overall}
\end{aligned}
\end{equation}
% \normalsize
where $L_{det}$ represents the detection loss function in the standard Faster-RCNN, including the classification and bounding boxes regression loss for the RPN and the final task. Note that we do not include any trade-off weight for each loss in $L_{ddf}$, which avoids the time-consuming process of hyperparameter fine-tuning. {\color{black}{In addition, the overall DDF framework is optimized in an end-to-end manner. The GRL layer between $D_{glb}$ and $E_{s}$ was employed when optimizing Equations~\ref{equ-loss-di} and~\ref{equ-loss-tri}.}}

During inference, the target images directly pass through the vanilla Faster-RCNN with off-the-shelf weights for object detection. For evaluation, we report the mean average precision (mAP) with a threshold of $0.5$. Our experiment is implemented with PyTorch~\cite{paszke2017automatic} on one NVIDIA GeForce 1080Ti GPU.

\begin{table*}[!htb]
\centering
\caption{Quantitative comparison with state-of-the-art methods on the Cityscapes $\to$ Foggy Cityscapes experiment. No DA represents the Faster-RCNN trained with the source images and directly tested on the target images, without any domain adaptation. For each comparison method, mAP$\star$ and gain represent its result under ``No DA" setting setting and the corresponding improvement achieved via adaptation, respectively. Oracle denotes the fully supervised Faster-RCNN trained on the target domain.}
% \resizebox{0.8\linewidth}{!}{%
\resizebox*{!}{0.30\textheight}{%
\begin{tabular}{l|l l l l l l l l |l|l|l}
\hline
Methods  & person & rider  & car & trunk  & bus& train & moto & bicycle & mAP & mAP$\star$ & gain \\
\hline
\hline
 & \multicolumn{8}{c|}{Backbone: VGG-16} & & &\\
\hline
No DA  & $23.2$ & $27.2$  & $32.8$ & $13.0$ & $23.5$ & $9.3 $& $12.9 $ & $25.3$ & $20.9$ & $20.9$ & $-$ \\
\hline
DAF~\cite{chen2018domain}  & $25.0$ & $31.0$  & $40.5$ & $22.1$ & $35.3$ & $20.2 $& $20.0 $ & $27.1$ & $27.6$ & $18.8$ & $8.8$\\
\hline
MAF~\cite{he2019multi}  & $28.2$ & $39.5$  & $43.9$ & $23.8$ & $39.9$ & $33.3 $& $29.2 $ & $33.9$ & $34.0$ & $18.8$ & $15.2$\\
\hline
SWDA~\cite{saito2019strong}  & $36.2$ & $35.3$  & $43.5$ & $30.0$ & $29.9$ & $ 42.3 $& $32.6 $ & $24.5$ & $34.3$ & $27.8$ & $6.5$\\
\hline
CFA~\cite{hsu2020every} & $41.9$ & $38.7$  & $\textbf{56.7}$ & $22.6$ & $41.5$ & $26.8 $& $24.6 $ & $35.5$ & $36.0$ & $18.8$ & $17.2$\\
\hline
CDN~\cite{su2020adapting}  & $35.8$ & $45.7$  & $50.9$ & $30.1 $ & $42.5$ & $29.8 $& $30.8 $ & $36.5$ & $36.6$ & $26.1$ & $10.5$\\
\hline
PD~\cite{wu2021instance}   & $33.1$ & $43.4$  & $49.6$ & $22.0$ & $45.8$ & $32.0 $& $29.6$ & $37.1$ & $36.6$ & $22.8$ & $13.8$\\
\hline
CDTD~\cite{shen2020cdtd} & $31.6$ & $44.0$  & $44.8$ & $30.4$ & $41.8$ & $40.7 $& $33.6 $ & $36.2$ & $37.9$ & $20.3$ & $17.6$\\
\hline
C2F~\cite{zheng2020cross}   & $\textbf{43.2}$ & $37.4$  & $52.1$ & $\textbf{34.7}$ & $34.0$ & $\textbf{46.9} $& $29.9 $ & $30.8$ & $38.6$ & $20.8$ & $17.8$\\
\hline
{\color{black}{FAA-SW~\cite{huang2021rda} }} & $39.5$ & $41.3$  & $47.0$ & $34.5$ & $39.3$ & $44.0  $& $31.9 $ & $28.4$ & $38.3$ & $22.0$ & $16.3$\\
\hline
MCAR~\cite{zhao2020adaptive}  & $32.0$ & $42.1$  & $43.9$ & $31.3$ & $44.1$ & $43.4 $& $37.4 $ & $36.6$ & $38.8$ & $23.4$ & $15.4$\\
\hline
DDF (ours)  & $37.2$ & $46.3 $  & $51.9$ & $24.7$ & $ 43.9 $ & $34.2 $& $33.5 $ & $\textbf{40.8}$ & $39.1$ & $20.9$ & $\textbf{18.2}$\\
\hline
Prior-DA~\cite{sindagi2020prior}   & $36.4$ & $\textbf{47.3}$  & $51.7$ & $22.8$ & $47.6$ & $34.1 $& $36.0$ & $38.7$ & $39.3$ & $24.4$ & $14.9$\\
\hline

Oracle  & $37.0$ & $47.2$  & $55.7$ & $31.1$ & $\textbf{54.4}$ & $29.6 $& $\textbf{38.5} $ & $40.2$ & $41.7$ & $20.9$ & $20.8$\\
\hline
\hline
 & \multicolumn{8}{c|}{{\color{black}{Backbone: ResNet-50}}} & & &\\
 \hline
 {\color{black}{No DA}}  & $28.7$ & $33.7$  & $36.7$ & $18.3$ & $30.1$ & $16.8 $& $21.7 $ & $27.3$ & $26.7$ & $-$ & $-$ \\
\hline
{\color{black}{MTOR~\cite{cai2019exploring}}}  & $30.6$ & $41.4$  & $44.0$  & $21.9$ & $38.6 $& $40.6$ & $28.3$ & $35.6$  & $35.1$ & $26.9$ & $8.2$\\
\hline
{\color{black}{GPA~\cite{xu2020cross}}}   & $32.9$ & $\textbf{46.7}$  & $54.1$ & $24.7$ & $45.7$ & $41.1 $& $\textbf{32.4}$ & $38.7$ & $39.5$ & $26.9$ & $12.6$\\
\hline
{\color{black}{DIDN~\cite{lin2021domain}}}   & $\textbf{38.3}$ & $44.4$  & $51.8$ & $28.7$ & $\textbf{53.3}$ & $34.7$& $\textbf{32.4}$ & $\textbf{40.4} $ & $40.5$ & $26.9$ & $13.6$\\
\hline
{\color{black}{UaDAN~\cite{guan2021uncertainty} }}  & $36.5$ & $46.1$  & $53.6$ & $28.9$ & $49.4$ & $42.7$& $32.3$ & $38.9$ & $41.1$ & $26.9$ & $14.2$\\
\hline
{\color{black}{DDF (ours)}}  & $37.6$ & $45.5 $  & $\textbf{56.1}$ & $\textbf{30.7}$ & $ 50.4 $ & $\textbf{47.0} $& $31.1 $ & $39.8$ & $\textbf{42.3}$ & $26.7$ & $\textbf{15.6}$\\
\hline

\end{tabular}}
\label{table-cmp-c2f}
\end{table*}

\section{Experimental Details}

\subsection{Dataset Description}

Our extensive experiments are conducted on four public datasets including Cityscapes~\cite{cordts2016cityscapes}, Foggy Cityscapes~\cite{sakaridis2018semantic}, SIM10K~\cite{johnson2016driving}, and KITTI~\cite{geiger2012we}.

\textbf{CityScapes}~\cite{cordts2016cityscapes} dataset contains $5000$ images of the real urban scenes collected from $27$ cities in different seasons. In this work, the ground truth for UDA object detection is the tightest bounding boxes for the instance mask annotations, following the previous practices in~\cite{chen2018domain,he2019multi,saito2019strong,chen2020harmonizing}. The $5000$ total images in the original dataset are officially split into training, validation, and testing sets with $2975$, $500$, and $1525$ images, respectively.

\textbf{Foggy CityScapes}~\cite{sakaridis2018semantic} dataset simulates foggy scenes in different intensity levels on the CityScapes dataset. Following the previous methods, we employed the images with the highest foggy intensity level. The data split and annotations of the Foggy Cityscapes dataset are the same as the Cityscapes dataset.

\textbf{SIM10K}~\cite{johnson2016driving} dataset contains $10000$ simulated images synthesized by the Grand Theft Auto (GTA) engine, with around $60K$ bounding box annotations of car instances. 

\textbf{KITTI}~\cite{geiger2012we} dataset contains $14999$ real autonomous driving images obtained from a mid-sized city, with around $80K$ bounding boxes annotations for object detection study. Following previous works~\cite{chen2018domain,he2019multi,he2020domain}, we employed the training set with $7481$ images for all experiments.

\subsection{Comparison Experiments}

\subsubsection{Adapting from the Normal to Foggy Weather}

Enhancing the model generalization ability under different weather conditions is crucial for automatic driving. In this section, we evaluate the model adaptation capability from the urban scene images captured in normal weather to the foggy ones, under the Cityscapes $\to$ Foggy Cityscapes setting. The training set with $2975$ images from the Cityscapes and Foggy Cityscapes datasets are employed as the source and target domains, respectively, to train the models. For testing, the models are validated on the validation set of the Foggy Cityscapes dataset with $500$ images. Under the same data split, our method is directly compared with the state-of-the-art methods, as shown in Table~\ref{table-cmp-c2f}. {\color{black}{In addition, we notice that the performance of the vanilla Faster-RCNN without adaptation is different in each comparison method. To this end, we first report the detection performance without adaptation for each method, which is denoted by the $mAP\star$ and directly copied from the original paper. Next, we report the performance gain of each method ($gain = mAP - mAP\star$) for a fair comparison, as suggested in~\cite{zheng2020cross}. Note that the performance gain can also be regarded as the adaptation ability for each cross-domain object detection method.}}

As illustrated in Table~\ref{table-cmp-c2f}, our DDF method has achieved the highest performance gain to the Faster RCNN without UDA. Compared with the Progressive Disentanglement (PD)~\cite{wu2021instance} and Conditional Domain Normalization (CDN)~\cite{su2020adapting} methods, which tackle the UDA object detection via feature disentanglement, our method has a better detection performance among most categories (7 out of 8). For the global-level disentanglement in PD~\cite{wu2021instance}, the calculation for the Mutual Information is suboptimal for decomposing the features in the UDA object detection architectures. In addition, its instance-level feature disentanglement requires reconstruction task learning, which introduces a decoder with auxiliary parameters. On the other hand, our GTD can avoid the influence from the batch size issue, and the ISD module requires no extra parameters, which further demonstrates the superiority and efficiency of our DDF method. {\color{black}{During the inference process of CDN~\cite{su2020adapting}, the target features for detection predictions contain the domain-specific information from both the source and target domains. Therefore, the detection model trained on the source data still suffers from domain bias, which would lead to suboptimal adaptation performance for the target data. By feature disentanglement for both the source and target domains, our DDF method is able to alleviate the influence of the target-specific factors during testing and achieve better adaptation performance.}}

Besides, we notice that the Coarse-to-Fine Feature Adaptation (C2F) method~\cite{zheng2020cross} has achieved the best performance under a few categories. Leveraging prototype alignment for the instance-level features, the C2F has a better object classification ability. However, our method can still achieve better overall performance by decomposing the irrelevant factors that cause domain shift. Although the mAP for Prior-DA~\cite{sindagi2020prior} has outperformed our DDF method, their mAP$\star$ is still inferior to ours. MAP$\star$ indicates the improvement of the proposed UDA method over the vanilla Faster R-CNN without adaption, which can alleviate the influence of the various implementation conditions for the comparison methods and therefore brings fairer comparison. Moreover, we notice that our DDF even outperforms the oracle fully supervised Faster-RCNN over two categories, which further demonstrates the feature adaptation ability of our DDF method. 

{\color{black}{
To further evaluate the effectiveness of our proposed method with various backbone models, we have also incorporated our DDF with Faster R-CNN using the ResNet50 backbone and compared with the state-of-the-art UDA object detection methods using the same backbone~\cite{cai2019exploring,xu2020cross,guan2021uncertainty,lin2021domain}. As indicated in Table~\ref{table-cmp-c2f}, our DDF method has achieved state-of-the-art detection and adaptation performance, which further demonstrates the robustness of the DDF method under different backbone models. Compared to DIDN~\cite{lin2021domain} on the disentanglement-based cross-domain object detection via feature reconstruction, our DDF method has shown superior adaptation performance by decomposing the features based on global-level triplet optimization and local-level similarity learning.
}}

\subsubsection{Adaptation from the Synthetic to Real Scene}

\begin{table}[!htb]
\centering
\caption{Quantitative comparison with state-of-the-art methods on the SIM10K $\to$ Cityscapes experiment. }
\resizebox{0.20\textheight}{!}{%
\begin{tabular}{l|l|l|l}
\hline
Methods  & mAP  & mAP$\star$ &gain\\
\hline
\hline
No DA  & $34.0$ & $34.0$ & $-$ \\
\hline
Baseline  & $39.5$ & $34.0$ & $5.5$ \\
\hline
DAF ~\cite{chen2018domain}  & $39.0$ & $30.1$ & $8.9$\\
\hline
SWDA~\cite{saito2019strong}  & $40.1$ & $34.6$ & $7.7$\\
\hline
MAF~\cite{he2019multi}  & $41.1$ & $30.1$ & $\textbf{11.0}$\\
\hline
SCDA~\cite{zhu2019adapting}  & $43.0$ & $34.0$ & $9.0$\\
\hline
ATF~\cite{he2020domain}  & $42.8$ & $34.6$ & $8.2$\\
\hline
HTCN~\cite{chen2020harmonizing}  & $42.5$ & $34.6$ & $7.9$\\
\hline
CDTD~\cite{shen2020cdtd}  & $42.6$ & $34.6$ & $8.0$ \\
\hline
{\color{black}{UMT~\cite{deng2021unbiased} }} & $43.1$ & $34.3$ & $8.8$ \\
\hline
C2F~\cite{zheng2020cross}  & $43.8$ & $35.0$ & $8.8$ \\
\hline
DDF (ours)  & $\textbf{44.3} $ & $34.0$ & $10.3$\\
\hline

\end{tabular}}
\label{table-cmp-s2c}
\end{table}

Learning from synthetic images is essential for object detection since it alleviates the extensive costs in acquiring the bounding and category annotations for each instance. In this section, we study the UDA object detection for the car object from the synthesized dataset to the real one, i.e., the adaptation from the SIM10K to Cityscapes dataset. During training, the whole SIM10K dataset is employed as the source domain, and the Cityscapes training set with $2975$ images as the target domain. The detailed results are shown in Table~\ref{table-cmp-s2c}. By outperforming the state-of-the-art methods such as HTCN~\cite{chen2020harmonizing} and UMT~\cite{deng2021unbiased}, the superiority of our proposed DDF method is further demonstrated. Among all the comparison methods, we notice that MAF~\cite{he2019multi} has a slightly better performance gain than ours. By outperforming MAF under other settings (Table~\ref{table-cmp-c2f}, and~\ref{table-cmp-kc-ck}), our DDF method is still more effective.

\subsubsection{Adaptation between Different Cameras}

\begin{table}[!htb]
\centering
\caption{Quantitative comparison with state-of-the-art methods on the UDA object detection tasks between the KITTI and Cityscapes datasets. $K \to C$ represents adaptation from the KITTI to the Cityscapes dataset, and vice versa.}
\resizebox{0.26\textheight}{!}{%
\begin{tabular}{l|l|l|l|l}
\hline
Methods  & \multicolumn{3}{c|}{$K \to C$} & $C \to K$ \\
\hline
\hline
Method  & mAP & mAP$\star$ & gain  & mAP \\
\hline
No DA  & $32.6$ & $32.6$ & $-$ & $53.5$ \\
\hline
Baseline  & $40.8$ & $32.6$ & $8.2$ & $68.6$ \\
\hline
DAF ~\cite{chen2018domain}  & $38.5$ & $30.2$ & $8.3$ & $64.1$ \\
\hline
SCDA~\cite{zhu2019adapting}   & $42.5$ & $37.4$ & $5.1$ & $-$ \\
\hline
MAF~\cite{he2019multi}  & $41.0$& $30.2$ & $10.8$  & $72.1$ \\
\hline
CDN~\cite{su2020adapting}  & $44.9$ & $37.1$ & $7.8$& $-$\\
\hline
{\color{black}{DSS~\cite{wang2021domain}}}  & $42.7$ & $34.6$ & $8.1$& $-$\\
\hline
{\color{black}{MeGA-CDA~\cite{vs2021mega}}} & $43.0$ & $30.2$ & $12.8$ & $\textbf{75.5}$\\
\hline
DDF (ours)  & $\textbf{46.0}$  & $32.6$ & $\textbf{13.4}$& $75.0$\\
\hline

\end{tabular}}
\label{table-cmp-kc-ck}
\end{table}

Due to a large diversity of the hardware devices, there exists a domain shift between two datasets captured through various cameras. In this section, we conduct a UDA car object detection experiment between the KITTI and Cityscapes datasets. For the adaptation from the KITTI to Cityscapes, the whole KITTI dataset is employed as a source domain, and the training set for the Cityscapes dataset is adopted as the target domain. When adapting from the Cityscapes to KITTI, the training set of the Cityscapes is employed as the source domain, and the whole KITTI dataset is used as the target domain. The results are presented in Table~\ref{table-cmp-kc-ck}. Under the KITTI $\to$ Cityscape setting, our method achieves the best performance, in terms of both the overall mAP and the performance gain. When adapting from Cityscapes to KITTI, we observe that the results for the ``No DA'' reported in all the other comparison methods~\cite{chen2018domain,he2019multi,he2020domain,vs2021mega} are the same. To reach a fair comparison, we directly report the comparisons on the overall mAP for each method, where our DDF also achieves competitive performance. 

\begin{figure*}[htbp]
\centering
\includegraphics[width=0.9\textwidth]{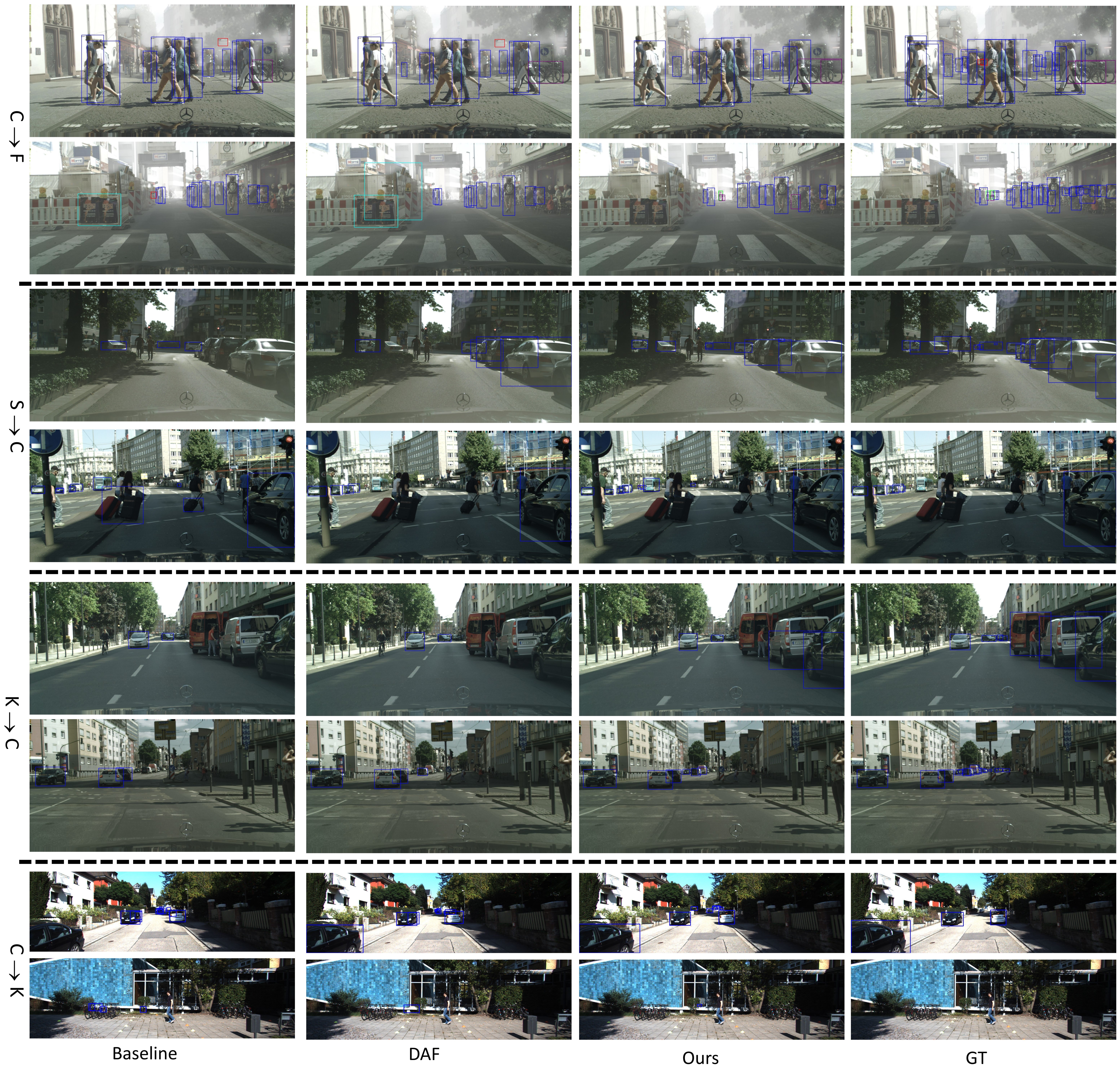}

\caption{{\color{black}{Visual comparison of results achieved under various UDA settings: C $\to$ F : CityScapes $\to$ Foggy CityScapes, S $\to$ C : Sim10K $\to$ CityScapes, K $\to$ C : KITTI $\to$ CityScapes, and C $\to$ K : CityScapes $\to$ KITTI. The bounding box predictions are overlapped on the original images.~\textbf{(Best viewed in color and zoom in)}}} }
\label{fig-vis-det-all}
\end{figure*}

{\color{black}{

We have also presented visual comparisons in Figure~\ref{fig-vis-det-all}, where the Baseline is the basic adaptation method without disentanglement, and DAF is a state-of-the-art UDA object detection method~\cite{chen2018domain}. It can be seen that our disentanglement-based method generated more true-positive predictions as well as fewer false predictions, further indicating the advantage of our method.

}}

\begin{figure*}[htbp]
\centering
\includegraphics[width=0.9\textwidth]{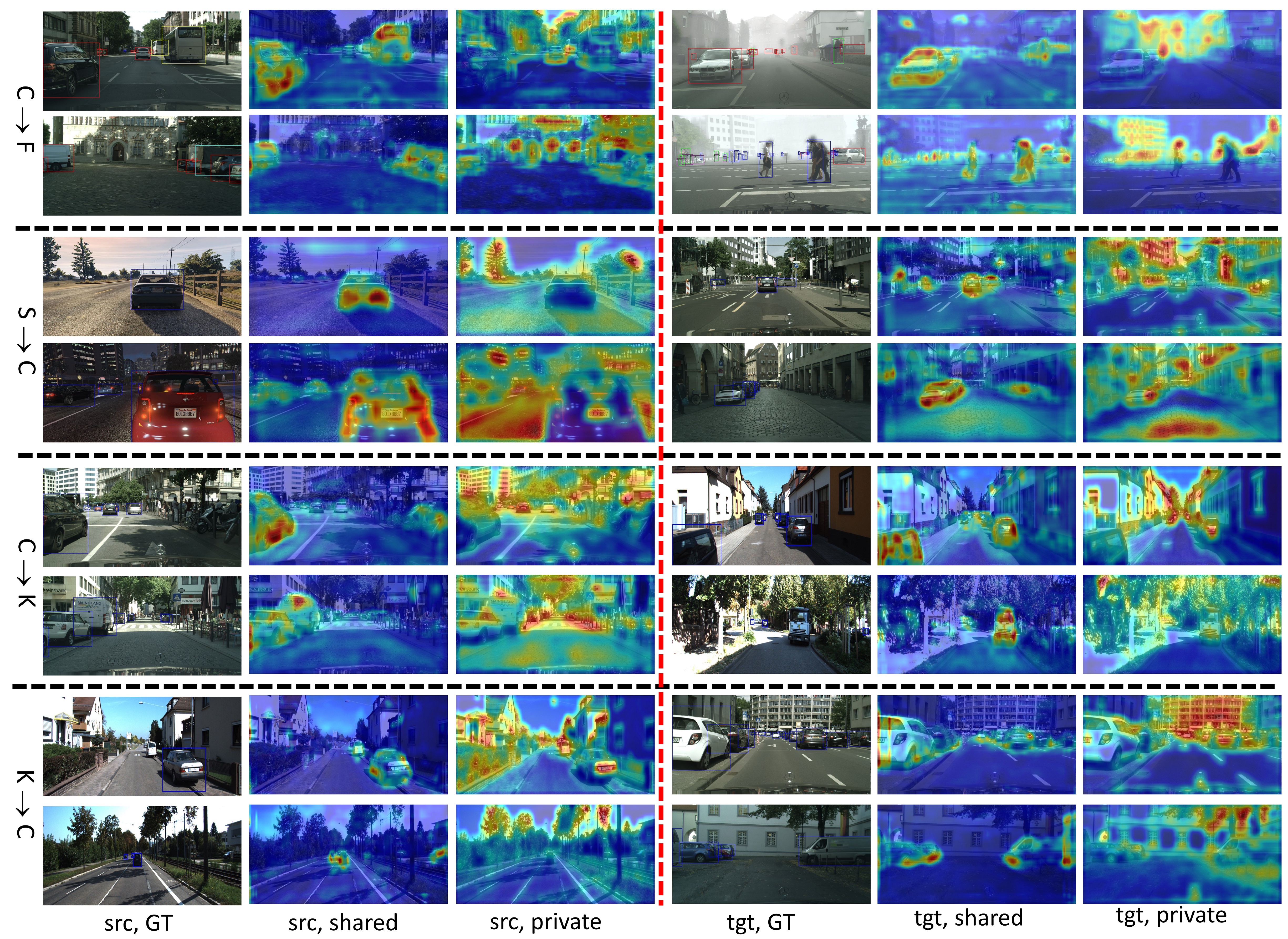}

\caption{Qualitative analysis on the disentanglement effectiveness of our DDF method under all UDA settings: C $\to$ F : CityScapes $\to$ Foggy CityScapes, S $\to$ C : Sim10K $\to$ CityScapes, K $\to$ C : KITTI $\to$ CityScapes, and C $\to$ K : CityScapes $\to$ KITTI. The left three columns are the results on the source domain (src), and the right three are on the target domain (tgt). All the features and labels are overlapped on the original images.~\textbf{(Best viewed in color and zoom in)} }
\label{fig-vis-feat}
\end{figure*}

\subsection{Ablation Studies}

In this section, we conduct an ablation analysis of the effectiveness of the proposed modules under the Cityscapes $\to$ Foggy Cityscapes setting. The results are shown in Table~\ref{table-abl-s2c}, where ``No DA'' represents the Faster-RCNN without domain adaptation and Baseline is the Faster-RCNN with global-level feature alignment ($L_{di}$ in Equation~\ref{equ-loss-di}). The w / o ISD method is implemented by removing $L_{ISD}$ in Equation~\ref{equ-loss-overall}, which can also be treated as Baseline + GTD. Similarly, the  w / o GTD method is equivalent to Baseline + ISD that neglects $L_{GTD}$ in Equation~\ref{equ-loss-overall}. Apart from the ablated modules, all the methods have the same implementation details as in Section~\ref{sec-imp}.

\begin{table*}[!htb]
\centering
\caption{Ablation studies for the GTD and ISD modules on the CityScapes $\to$ Foggy Cityscapes experiment.}
\resizebox{0.65\linewidth}{!}{%
\begin{tabular}{l|l l l l l l l l |l}
\hline
Method   & person & rider  & car & trunk  & bus& train & moto & bicycle & mAP \\
\hline
No DA  & $23.2$ & $27.2$  & $32.8$ & $13.0$ & $23.5$ & $9.3 $& $12.9 $ & $25.3$ & $20.9$  \\
\hline
Baseline   & $25.8$ & $32.8$  & $46.1$  & $21.4$ & $36.9$& $24.7 $ & $17.9$ & $25.7 $& $28.9$ \\
\hline
w / o GTD &   $33.2$ & $41.0$  & $51.0$ & $\textbf{30.0}$ & $43.4$ & $24.0 $& $25.8 $ & $35.1$ & $35.4$ \\ 
\hline
{\color{black}{w / o $L_{ds}$}} &   $36.6$ & $\textbf{46.9}$  & $55.2$ & $28.4$ & $46.3$ & $26.9 $& $28.4 $ & $38.7$ & $38.4$ \\ 
\hline
w / o $L_{tri}$   &   $36.8$ & $44.7$  & $\textbf{55.4}$ & $26.5$ & $\textbf{45.1}$ & $27.1 $& $28.8 $ & $38.7$ & $37.9$ \\ 
\hline
w / o ISD   & $35.0$ & $44.4$  & $51.0$ & $23.4$ & $42.0 $& $22.3 $ & $30.0$ & $38.2$  & $35.8$\\
\hline
{\color{black}{w / o $L_{ISD-intra}$ }} &   $36.0$ & $42.2$  & $54.0$ & $\textbf{27.2}$ & $42.5$ & $25.1 $& $32.2 $ & $37.4$ & $37.1$ \\ 
\hline
w / o $L_{ISD-inter}$  &   $35.7$ & $44.0$  & $51.5$ & $25.6$ & $40.5$ & $\textbf{36.0} $& $28.6 $ & $38.5$ & $37.5$ \\ 
\hline
DDF  & $\textbf{37.2}$ & $46.3 $  & $51.9$ & $24.7$ & $ 43.9 $ & $34.2 $& $\textbf{33.5} $ & $\textbf{40.8}$ & $\textbf{39.1}$ \\
\hline

\end{tabular}}
\label{table-abl-s2c}
\end{table*}

Aligning the features at the global level, the Baseline outperforms the ``No DA'' method among all categories. To further strengthen the model adaptation ability at the global and local levels, the GTD and ISD modules are proposed to bring consistent performance gains under most categories except for ``trunk'' and ``train''. As two types of large-scale vehicles, parts of the trains and trunks are more likely to be covered by the fog and thus become obscure and vague for the perception system, compared to the objects from the rest categories. To this end, detecting these two categories under the foggy situation is challenging, which is also proved by the inferior performance achieved by the fully supervised benchmark shown in Table~\ref{table-cmp-c2f}.

Although solely utilizing the GTD or ISD models incurs a slight performance drop on the ``train'' class, integrating them can still improve the performance significantly (about $10\%$). This indicates the cross-domain detection for the ``train'' class is challenging and requires feature disentanglement at both global and local stages. In addition, we notice that the DDF incorporating ISD and GTD degrades the performance than only using ISD. However, by achieving the best overall detection performance, incorporating GTG with ISD in DDF method is still demonstrated to be superior.

{\color{black}{

Moreover, we also include ablation studies on the inter-domain ISD ($L_{ISD-inter}$ in Equation~\ref{equ-ins-inter}), intra-domain ISD ($L_{ISD-intra}$ in Equation~\ref{equ-ins-intra}), global-level triplet optimization mechanism ($L_{tri}$ in Equation~\ref{equ-loss-tri}), and the global-level domain-specific classification learning ($L_{ds}$ in Equation~\ref{equ-loss-ds}). The results show performance drop by removing each module, hence demonstrating that all modules are useful. For the global-level GTD module, we notice the triplet optimization module has a stronger decomposition ability than the domain-specific classifier, as indicated by the larger performance drop when ablated. At the local level, the inter-domain ISD works slightly better than the intra-domain one.

}}

\begin{table}[!htb]
\centering
\caption{Quantitative evaluation on the domain distance under the Cityscapes $\to$ Foggy Cityscapes setting. G and I represent the features at the global and local levels, respectively.}
\resizebox{0.22\textheight}{!}{%
\begin{tabular}{l|l|l|l|l}
\hline
  & \multicolumn{2}{c|}{PAD$\downarrow$}  & \multicolumn{2}{c}{EMD$\downarrow$} \\
\hline
  & G & I  & G & I \\
\hline
No DA & $1.98$ & $1.69$ & $6.37$ & $10.46$ \\
\hline
Baseline  & $1.51$ & $0.53$ & $1.83$ & $10.34$\\
\hline
Ours & $\textbf{0.75}$ & $\textbf{0.39}$ & $\textbf{1.38}$ & $\textbf{9.64}$\\
\hline
\end{tabular}}
\label{table-cmp-distance}
\end{table}

\subsection{Disentanglement Evidence Analysis}

In this section, we first perform a qualitative analysis on the effectiveness of our disentanglement-based DDF method. For the No DA and Baseline, the visualized features are the outputs of the backbone network. For our DDF method, we visualize the domain-shared features as well as the domain-private features at the global level. All the features are firstly averaged along their channel dimensions and then normalized to range $[0,255]$. Finally, the features are resized and overlapped with the original images for visualization. {\color{black}{Note that we only visualize the image-level features. The instance-level features $I_{sha.}^s$, $I_{sha.}^t$, $I_{pri}^s$, and $I_{pri}^t$, on the other hand, are vectors instead of tensors. Therefore, they do not have height and width and cannot be resized and overlaid on the original images for visualization.
}}

As shown in Figures~\ref{fig-vis-feat} and~\ref{fig-disent-c2f-intro}, the domain-shared features from our DDF at the global stages particularly focus on the instance objects which are crucial for detection task learning. However, the domain-shared features for the Baseline still focus on auxiliary background components, instead of particularly on the foreground instances. Therefore, in the Baseline method, the global-level domain-specific factors are introduced into the detection task learning, which results in the feature entanglement issue and further degrades the model adaptation performance. In addition, we observe the domain-private features for our DDF particularly concentrate on the information specific to its belonging domain, such as the components that reflect the weather conditions. This further demonstrates our $E_{p}$ and $E_{s}$ successfully extract the private and shared features for the given domains as expected.

To quantitatively evaluate the disentanglement ability of our DDF method, we calculate the feature distribution discrepancy of the domains under the Cityscape $\to$ Foggy Cityscape setting at both the global and local stages. At the global level, the feature distributions are acquired by average pooling applied on the outputs of the network backbone. For the local-level features, we randomly select $100$ foreground object features from each category in each domain. To measure the domain distance, Proxy $\mathcal{A}$-distance (PAD)~\cite{ben2007analysis} and Earth Movers Distance (EMD)~\cite{rubner2000earth} are employed. As shown in Table~\ref{table-cmp-distance}, our method achieves a shorter feature distribution distance than the Baseline at both the global and local levels. Given a better representation of the domain-invariant features and a lower cross-domain feature discrepancy, our DDF method is proved to be effective in improving the Baseline via feature disentanglement.

\subsection{{\color{black}{Model Design Selections}}}

{\color{black}{
In this section, we discuss the effectiveness of our DDF methods under different model design choices, including directly considering the similarity between the local-level shared cross-domain features, and introducing the adversarial learning strategies at the local level. The experiments are conducted under the Cityscape $\to$ Foggy Cityscape setting and the results are shown in Table~\ref{table-s2c-selections}.
}}

\begin{table}[!htb]
\centering
\caption{{\color{black}{The DDF method with different model designs on the CityScapes $\to$ Foggy Cityscapes experiment. `ins-simmax' represents the DDF with the direct similarity maximization between the cross-domain shared features at the local level. `ins-td ' indicates the DDF employing triplet disentanglement module at the local level.}}}
\resizebox{0.95\linewidth}{!}{%
\begin{tabular}{l|l l l l l l l l |l}
\hline
Method   & per & rid  & car & tru  & bus& tra & moto & bic & mAP \\
\hline
ins-simmax  & $32.4$ & $36.6 $  & $43.5$ & $24.1$ & $ 40.7 $ & $16.5 $& $22.4 $ & $32.2$ & $31.1$ \\
\hline
ins-td  & $37.6$ & $46.2 $  & $51.9$ & $23.4$ & $ 43.8 $ & $23.7 $& $34.8 $ & $39.0$ & $37.6$ \\
\hline
OG DDF  & $37.2$ & $46.3 $  & $51.9$ & $24.7$ & $ 43.9 $ & $34.2 $& $33.5 $ & $40.8$ & $39.1$ \\
\hline

\end{tabular}}
\label{table-s2c-selections}
\end{table}

{\color{black}{
First, we include a similarity maximization mechanism for the local-level domain-shared features $I_{sha.}^{s}$ and $I_{sha.}^{t}$, by inducing the cosine similarity between $I_{sha.}^{s}$ and $I_{sha.}^{t}$ to be $1$. This selection is referred to as `ins-simmax' in Table~\ref{table-s2c-selections}. By directly maximizing the similarity between the cross-domain shared features, the model performance drops significantly under many categories. One of the major reasons is that the cross-domain local-level features are ROIs under several categories, and directly narrowing the distribution gap between them incurs misalignment across different classes. 
}}

{\color{black}{
Next, we conduct an experiment replacing the similarity-based ISD module of the DDF method at the local level with the same disentanglement strategies as the global level, i.e., using the adversarial learning strategy and triplet optimization. Specifically, the optimization strategies in Equations~\ref{equ-loss-di},~\ref{equ-loss-ds}, and~\ref{equ-loss-tri} are adopted into $I_{sha.}^{s}$, $I_{sha.}^{t}$, $I_{pri.}^{s}$ and $I_{pri.}^{t}$ at the local level. The overall method is referred to as `ins-td' in Table~\ref{table-s2c-selections}, which achieves less competitive performance than the DDF method with similarity-based feature disentanglement at the local level. Due to the lack of annotations for the target images during training, the instance-level target features might not be accurate without the detection task learning, especially at the early training stage. Therefore, we think the inferior performance of the `ins-td' method results from the influence of the unstable training process of the adversarial learning strategy. In addition, a similar phenomenon was also observed in~\cite{saito2019strong} previously: including an adversarial domain discriminator for the local features in a UDA object detection model degrades the overall performance.
}}

\subsection{Computational Complexity Analysis}

In this section, we conduct a computational complexity analysis for each principle module (GTD and ISD) in our proposed DDF method. The computational cost for the ablation studies under the $C \to F$ settings in Table~\ref{table-abl-s2c} is as follows (number of parameters, and training time per iteration): 1) Baseline: $140M$, $0.34s/iter$; 2) Baseline + GTD: $145M$, $0.39s/iter$; 3) Baseline + ISD: $145M$, $0.51s/iter$; and 4) DDF: $145M$, $0.54s/iter$. For the model size, the auxiliary cost for both ISD and GTD modules is negligible. Although our ISD module brings extra training time, we think it is acceptable given the performance gain.

\section{Conclusion}

In this work, we present a novel Domain Disentanglement Faster-RCNN (DDF) method for cross-domain object detection via feature disentanglement. Specifically, we propose a GTD module to decompose the shared and private features within each domain at the global stage, as well as fitting the model batch size restrictions. At the instance level, an ISD module based on inter- and intra-domain similarity optimization is proposed to facilitate the feature disentanglement. Extensive experiments demonstrate the superiority of our DDF method by outperforming state-of-the-art methods on several benchmark UDA object detection tasks. Additionally, the qualitative and quantitative analysis on the disentanglement evidence further indicates the effectiveness of our method on decomposing the domain-specific factors and eliminating the domain-bias from them. In a larger perspective, the feature entanglement issues and the restrictions on model batch size are not limited to cross-domain object detection. With the promising performance on cross-domain object detection, the DDF method can also be extended to other cross-domain object analysis tasks (e.g., segmentation, and tracing) in future work.

{\small

\bibliographystyle{IEEEtran}
\bibliography{IEEEabrv, ref}

}

\end{document}